\documentclass{article}

\usepackage{wrapfig}
\usepackage[utf8]{inputenc} 
\usepackage[T1]{fontenc}    
\usepackage{hyperref}       
\usepackage{url}            
\usepackage{booktabs}       
\usepackage{amsfonts}       
\usepackage{nicefrac}       
\usepackage{microtype}      
\usepackage{graphicx}
\usepackage{subfigure}
\usepackage{amsfonts}
\usepackage{amsmath}
\usepackage{multirow}
\usepackage{amsmath}
\usepackage{amssymb}
\usepackage{amsthm}
\usepackage{algorithm}
\usepackage{algorithmic}
\usepackage{color}
\usepackage[left=2.5cm, right=2.5cm, top=2cm]{geometry}

\newcommand{\mc}{\mathcal}
\newcommand{\mb}{\mathbf}
\newcommand{\bs}{\boldsymbol}
\newcommand{\mbb}{\mathbb}

\title{Missing Value Imputation Based on Deep Generative Models}
\author{Hongbao Zhang, Pengtao Xie, Eric Xing\\
  \small Petuum, Inc
}
\date{}

\begin{document}
\maketitle
\begin{abstract}
Missing values widely exist in many real-world datasets, which hinders the performing of advanced data analytics. Properly filling these missing values is crucial but challenging, especially when the missing rate is high. Many approaches have been proposed for missing value imputation (MVI), but they are mostly heuristics-based, lacking a principled foundation and do not perform satisfactorily in practice. In this paper, we propose a probabilistic framework based on deep generative models for MVI. Under this framework, imputing the missing entries amounts to seeking a fixed-point solution between two conditional distributions defined on the missing entries and latent variables respectively. These distributions are parameterized by deep neural networks (DNNs) which possess high approximation power and can capture the nonlinear relationships between missing entries and the observed values. The learning of weight parameters of DNNs is performed by maximizing an approximation of the log-likelihood of observed values. We conducted extensive evaluation on 13 datasets and compared with 11 baselines methods, where our methods largely outperforms the baselines.
\end{abstract}

\section{Introduction}

In real world datasets, missing values are pervasive, caused by incomplete measurement, data corruption, privacy concerns, and so on. For example, consider lab test values in a clinical setting: it is typical
that at a certain time point only a small subset of laboratory tests are examined, leaving the values of most tests missed. Such missingness incurs substantial difficulty for analyzing data and distilling insights therefrom. 

To address this issue, it is important to perform missing value imputation (MVI): guessing the missing entries based on the observed ones. By filling the missing entries correctly, one can obtain a complete picture of the data which facilitates the  extraction of more informed patterns. In addition, since most machine learning (ML) models require the inputs to be fixed-length feature vectors, MVI makes incomplete data ML-ready. In general, MVI is a challenging task, especially when the the missing rate is high. This is due to the following reasons. First, the relationship between observed values and missing values, which is the foundation of designing MVI methods, is unknown in most cases. Previous methods~\cite{DBLP:journals/corr/GondaraW17, Tran2017MissingMI} make simplistic assumptions such as the relationship is linear, which may not hold in practice. Second, the pattern of missingness is typically irregular: different data instances have different number of missing entries and the positions of these entries are different as well. Such irregularity hinders the design of MVI models. For example, suppose one wants to build a predictive model that maps the observed entries to missing entries. Mathematically, the inputs and outputs of such a model is difficult to be formulated since the missing entries of different examples have different positions.

A variety of methods have been proposed for MVI. While showing promising results, they have certain drawbacks. Several approaches~\cite{DBLP:journals/corr/GondaraW17, Tran2017MissingMI} use autoencoders for MVI. While empirically effective, they are heuristics-based and lack a principled foundation. Specifically, it is unclear what mathematical objective is defined w.r.t the missing values.
Some methods perform MVI based on matrix completion \cite{cai2010singular, mazumder2010spectral, keshavan2010matrix}, by assuming the underlying complete data matrix is low-rank. Such an assumption essentially posits that the relationship between observed entries and missing entries is linear. Hence these methods are unable to explore the nonlinear relationship, which are more common in practice. In many methods such as mean imputation, the missing entries are imputed at a population-level: for each feature, the same value (e.g., mean of all observed values of this feature) is used to impute all missing instances of this feature across different data examples. This is problematic since two examples that may be completely different are given the same value.   

In this paper, we aim at developing an MVI method that addresses the limitations of existing approaches. First, we formulate MVI as estimating the conditional distribution $p(\mc{Y}|\mc{X})$ of missing entries $\mc{Y}$ based on the observed values $\mc{X}$, and show that this distribution can be inferred by solving a fixed-point problem (FPP). Such a formulation enables our method to perform example-level imputation where the missing values of a data example is imputed based on the observed values of this example. This FPP involves two conditional distributions bridged by a latent variable. To capture the nonlinear relationship between $\mc{Y}$ and $\mc{X}$, we parameterized the two conditional distributions using deep neural networks (DNNs). DNNs are nonlinear functions that possess great approximation power. Hence our method can well capture the nonlinear relationship between observed and missing values. The design of these two distributions enables our method to adapt to any irregularity of positions and numbers of the missing values. The weight parameters of DNNs are estimated by maximizing the approximated log-likelihood of all observed data, via an expectation maximization (EM) algorithm. In the E step, MVI is performed by fixing the model parameters. In the M step, model parameters are updated based on the imputed missing values. Compared with previous approaches, our method possesses the following merits: (1) a well-grounded theoretical foundation; (2) ability of capturing nonlinear relationship between missing and observed values; (3) ability of performing instance-level imputation. We demonstrate the empirical effectiveness of our method on 13 datasets by comparing with 11 baselines.

The rest of this paper is organized as follows. In Section 2, we introduce related works. Section 3 and 4 present the methods and experimental results. 
Section 5 concludes the paper. 

\section{Related Work}
Dempster et al~\cite{dempster1977maximum} propose an EM algorithm for maximum likelihood estimation from incomplete data. Our method differs from this work in two-fold. First, we estimate the conditional distribution $p(\mc{Y}|\mc{X})$ by solving a fixed-point problem. Second, the imputation models of our methods are parameterized by deep neural networks. Several works \cite{cai2010singular,mazumder2010spectral,keshavan2010matrix} have studied missing value imputation from a matrix completion perspective, either via nuclear norm regularization~\cite{cai2010singular,mazumder2010spectral} or matrix factorization~\cite{keshavan2010matrix}. Buuren and Groothuis-Oudshoorn~\cite{buuren2010mice} propose to iteratively impute the values of each feature through regression on the values of remaining features. All these methods are inherently linear, which may not be sufficient to capture the nonlinear patterns between the observed and missing values. And these methods need to perform singular value decomposition in each iteration, which is computationally heavy. Deep learning models have been explored for MVI. Gondara and Wang~\cite{DBLP:journals/corr/GondaraW17} propose a multiple imputation model based on overcomplete deep denoising autoencoders~\cite{vincent2008extracting}. Tran el al.~\cite{Tran2017MissingMI} propose a cascaded residual autoencoder (CRA) to impute missing
modalities. By stacking residual autoencoders, CRA grows
iteratively to model the residual between the current prediction
and original data. While showing empirical effectiveness, these methods lack a theoretical foundation. Stekhoven and B{\"u}hlmann~\cite{stekhoven2011missforest} develop a nonparametric method to cope with the imputation of different types of variables simultaneously. Multiple imputation by chained equations (MICE)~\cite{buuren2010mice} generates imputations based on a set of imputation models, one for each variable with missing values. Similarly, these methods are heuristics-based. A theoretical understanding is missing.  

\section{Method}\label{sec:method}

In this section, we study how to perform missing value imputation based on deep generative models. Overall there are three types of missing mechanism: (1) missing completely at random (MCAR), where the propensity for a data item to be missing is completely random; (2) missing at random (MAR), where the propensity for a data item to be missing is not related to the missing data, but it is related to some of the observed data; (3) missing not at random (MNAR), where the data items are missing due to certain underlying mechanisms. In this work, we mainly focus on MNAR.

We assume there are $K$ features at the population level. For each individual data example $n$, it has $O^{(n)}$ ($0\le O^{(n)} \leq K$) observed features $\mc{X}^{(n)}=\{x_i\}_{i=1}^{O^{(n)}}$ and the rest $K-O^{(n)}$ features $\mc{Y}^{(n)}=\{y_i\}_{i=1}^{K-O^{(n)}}$ are missing. For different data examples, the missing features are different. We formulate the missing value imputation problem in a probabilistic framework: (1) treat observed and missing features as random variables; (2) given the observed features $\mc{X}^{(n)}$, infer the conditional distribution of the missing ones $p(\mc{Y}^{(n)}|\mc{X}^{(n)})$. We introduce a latent variable $\mb{z}^{(n)}\in \mathbb{R}^{J}$ to learn the latent representation of this data example and assume $\mc{X}^{(n)}$ and $\mc{Y}^{(n)}$ are conditionally independent given $\mb{z}^{(n)}$:
\begin{equation}
p(\mc{X}^{(n)},\mc{Y}^{(n)}|\mb{z}_n)=p(\mc{X}^{(n)}|\mb{z}_n)p(\mc{Y}^{(n)}|\mb{z}_n).
\end{equation}

\begin{figure}[t]
\begin{center}
\includegraphics[width=0.8\columnwidth]{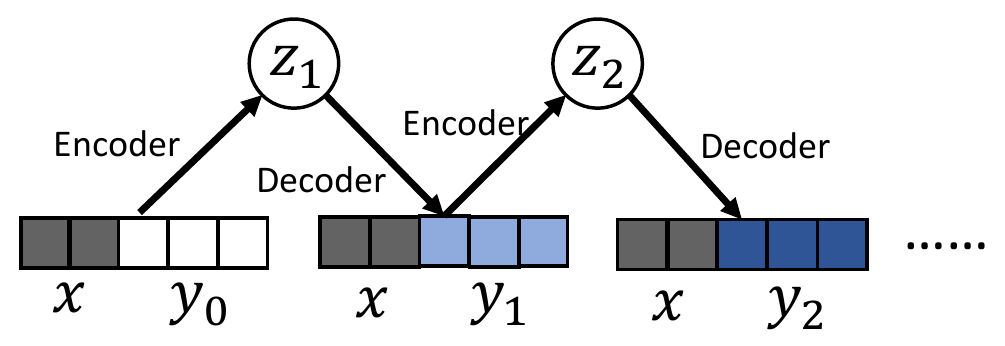}
\end{center}
\caption{\footnotesize An iterative process of imputing missing values.
}
\label{fig:ae_inference}
\end{figure}

In the sequel, we drop $n$ from the notations for simplicity. Based on $\mb{z}$, the conditional distribution $p(\mc{Y}^{(n)}|\mc{X}^{(n)})$ can be calculated in the following way:
\begin{equation}
\centering
\label{eq:y_x}
\begin{array}{l}
p(\mc{Y}|\mc{X})
=\int_{\mb{z}}p(\mc{Y},\mb{z}|\mc{X})
=\int_{\mb{z}}\frac{p(\mc{X},\mc{Y},\mb{z})}{p(\mc{X})}
=\int_{\mb{z}}\frac{p(\mc{X},\mc{Y}|\mb{z})p(\mb{z})}{p(\mc{X})}
=\int_{\mb{z}}\frac{p(\mc{Y}|\mb{z})p(\mc{X}|\mb{z})p(\mb{z})}{p(\mc{X})}
=\int_{\mb{z}}\frac{p(\mc{Y}|\mb{z})p(\mc{X},\mb{z})}{p(\mc{X})}\\
=\int_{\mb{z}}p(\mc{Y}|\mb{z})p(\mb{z}|\mc{X})
=\mathbb{E}_{p(\mb{z}|\mc{X})}[p(\mc{Y}|\mb{z})]
\approx p(\mc{Y}|\mathbb{E}[\mb{z}|\mc{X}])
\end{array}
\end{equation}
where the last step is first-order Taylor approximation of $\mathbb{E}_{p(\mb{z}|\mc{X})}[p(\mc{Y}|\mb{z})]$. An intuitive interpretation of this equation is that: given the observed features, we first infer the latent representation $\mbb{E}[\mb{z}|\mc{X}]$, then generate the missing features by inferring $p(\mc{Y}|\mbb{E}[\mb{z}|\mc{X}])$. To calculate $\mbb{E}[\mb{z}|\mc{X}]$, we need to infer $p(\mb{z}|\mc{X})$, which can be done in the following way:
\begin{equation}
\label{eq:z_x}
\begin{array}{l}
p(\mb{z}|\mc{X})
=\int_{\mc{Y}}p(\mb{z},\mc{Y}|\mc{X})
=\int_{\mc{Y}}\frac{p(\mb{z},\mc{Y},\mc{X})}{p(\mc{X})}
=\int_{\mc{Y}}\frac{p(\mb{z}|\mc{X},\mc{Y})p(\mc{X},\mc{Y})}{p(\mc{X})}
=\int_{\mc{Y}}p(\mb{z}|\mc{X},\mc{Y})p(\mc{Y}|\mc{X})\\
=\mathbb{E}_{p(\mc{Y}|\mc{X})}[p(\mb{z}|\mc{X},\mc{Y})]
\approx p(\mb{z}|\mc{X},\mathbb{E}[\mc{Y}|\mc{X}])]\\
\end{array}
\end{equation}
Similarly, the last step is obtained using first order Taylor approximation. As can be seen from Eq.(\ref{eq:y_x}) and Eq.(\ref{eq:z_x}), $p(\mc{Y}|\mc{X})$ and $p(\mb{z}|\mc{X})$ can be estimated using fixed-point iteration method which iteratively performs the following two steps until convergence: (1) given $p(\mc{Y}|\mc{X})$, substitute it into $p(\mb{z}|\mc{X},\mathbb{E}[\mc{Y}|\mc{X}])]$ as an approximation of $p(\mb{z}|\mc{X})$; (2) given $p(\mc{Y}|\mc{X})$, substitute it into $p(\mb{z}|\mc{X},\mathbb{E}[\mc{Y}|\mc{X}])]$ as an approximation of $p(\mb{z}|\mc{X})$. Figure ~\ref{fig:ae_inference} illustrates this process. We initialize the missing values $\mathbb{E}[\mc{Y}]$ with $y_0$, concatenate it with the observed values $x$, and feed the concatenation to $p(\mb{z}|\mc{X},\mathbb{E}[\mc{Y}|\mc{X}])$, which is called an encoder. Then we calculate $\mbb{E}[\mb{z}|\mc{X},\mathbb{E}[\mc{Y}|\mc{X}]]$ and feed it to $p(\mc{Y}|\mathbb{E}[\mb{z}|\mc{X}])$, which is called a decoder to get a refined estimation of the missing entries. This two steps iterates until convergence.

Next, we discuss how to parameterize $p(\mc{Y}|\mathbb{E}[\mb{z}|\mc{X}])$ and $p(\mb{z}|\mc{X},\mathbb{E}[\mc{Y}|\mc{X}])]$. Similar to variational autoencoder~\cite{kingma2013auto}, we use deep neural networks to perform such parameterization. To parameterize $p(\mc{Y}|\mathbb{E}[\mb{z}|\mc{X}])$, we first generate two $K$ dimensional vectors using deep networks (represented by $f(\cdot)$ and $g(\cdot)$): $\bs\mu=f(\mathbb{E}[\mb{z}|\mc{X}])$ and $\bs\sigma=g(\mathbb{E}[\mb{z}|\mc{X}])$. Then for each missing feature $y\in\mc{Y}$, let $i\in\{1,\cdots,K\}$ denote its index. If $y$ is a continuous variable, the distribution defined on $y$ is a univariate Gaussian $\mc{N}(y|\mu_i, \sigma_i)$ where $\mu_i$ and $\sigma_i$ are the $i$-th dimension of $\bs\mu$ and $\bs\sigma$. If $y$ is a binary variable, then the distribution is a Bernoulli distribution where $p(y=1)=1/(1+\exp(-\mu_i))$. Similarly, we use deep networks (represented by $h(\cdot)$ and $l(\cdot)$) to parameterize $p(\mb{z}|\mc{X},\mathbb{E}[\mc{Y}|\mc{X}])$ as a multivariate Gaussian distribution $\mc{N}(\mb{z}|h(\mc{X}\oplus \mathbb{E}[\mc{Y}|\mc{X}]), l(\mc{X}\oplus \mathbb{E}[\mc{Y}|\mc{X}]))$ and $\mc{X}\oplus \mathbb{E}[\mc{Y}|\mc{X}]$ denotes the concatenation of the observed features $\mc{X}$ and imputed features $\mathbb{E}[\mc{Y}|\mc{X}]$ (in the order of their indexes).

Next, we discuss how to learn the parameters of the deep networks. First, we show that the log-likelihood of the observed values can be approximated by a function of $p(\mc{Y}|\mc{X})$, $p(\mb{z}|\mc{X},\mbb{E}[\mc{Y}|\mc{X}])$, $p(\mc{X}|\mb{z})$, and $p(\mbb{E}[\mc{Y}|\mc{X}]|\mb{z})$, where $p(\mc{X}|\mb{z})$ and $p(\mbb{E}[\mc{Y}|\mc{X}]|\mb{z})$ are parameterized by the same neural networks $f(\cdot)$ and $g(\cdot)$ defined before.
\begin{equation}
\label{eq:vi_x_y}
\begin{array}{l}
\log p(\mc{X})\\
= \mbb{E}_{p(\mc{Y}|\mc{X})}[\log p(\mc{X},\mc{Y})]
-\mbb{E}_{p(\mc{Y}|\mc{X})}[\log p(\mc{Y}|\mc{X})]\\
\approx \log p(\mc{X},\mbb{E}[\mc{Y}|\mc{X}])
-\mbb{E}_{p(\mc{Y}|\mc{X})}[\log p(\mc{Y}|\mc{X})]\\
=\mbb{E}_{p(\mb{z}|\mc{X},\mbb{E}[\mc{Y}|\mc{X}])}[\log p(\mc{X},\mbb{E}[\mc{Y}|\mc{X}],\mb{z})]-\mbb{E}_{p(\mb{z}|\mc{X},\mbb{E}[\mc{Y}|\mc{X}])}[\log p(\mb{z}|\mc{X},\mbb{E}[\mc{Y}|\mc{X}])]
-\mbb{E}_{p(\mc{Y}|\mc{X})}[\log p(\mc{Y}|\mc{X})]\\
=\mbb{E}_{p(\mb{z}|\mc{X},\mbb{E}[\mc{Y}|\mc{X}])}[\log p(\mc{X}|\mb{z})]+\mbb{E}_{p(\mb{z}|\mc{X},\mbb{E}[\mc{Y}|\mc{X}])}[\log p(\mbb{E}[\mc{Y}|\mc{X}]|\mb{z})]+\mbb{E}_{p(\mb{z}|\mc{X},\mbb{E}[\mc{Y}|\mc{X}])}[\log p(\mb{z})]\\
\;\;\;-\mbb{E}_{p(\mb{z}|\mc{X},\mbb{E}[\mc{Y}|\mc{X}])}[\log p(\mb{z}|\mc{X},\mbb{E}[\mc{Y}|\mc{X}])]
-\mbb{E}_{p(\mc{Y}|\mc{X})}[\log p(\mc{Y}|\mc{X})]
\end{array}
\end{equation}
We then perform parameter learning by maximizing the approximation of $\log p(\mc{X})$ using an expectation maximization (EM) algorithm, which iteratively performs the following two steps until convergence. First, in the E(expectation) step, we fix the weight parameters of neural networks and infer $p(\mc{Y}|\mc{X})$. This is basically the missing value imputation task, which can be accomplished by executing the fixed-point iteration method between Eq.(\ref{eq:y_x}) and Eq.(\ref{eq:z_x}). The (fixed) weight parameters of neural networks are needed to calculate these two equations. In the (M)aximization step, we fix $p(\mc{Y}|\mc{X})$ (and therefore $\mbb{E}[\mc{Y}|\mc{X}]$), and estimate the model parameters. The estimation can be conducted using the stochastic
gradient variational Bayes algorithm~\cite{kingma2013auto}.

\section{Experiment}
We evaluated our method on various publicly available datasets, with various percentages of missing values and compared with 11 baseline methods both quantitatively and qualitatively.

\subsection{Datasets}
We use 13 datasets from the UCI repository: 10 for classification tasks and 3 for regression tasks. Table \ref{tab:data-dscription} shows the statistics of these datasets. The numbers of data examples vary from about 100 to over 3000 while the numbers of features vary from 4 to 60. These features are of various types, including ordinal, categorical, and continuous. Min-max normalization was used for feature normalization.

\begin{table}[t]
\centering
\caption{Statistics of datasets}
\label{tab:data-dscription}
\resizebox{\textwidth}{!}{
\begin{tabular}{l|cccc}
\toprule
          &\# Training & \# Testing &\# Features & Data types                        \\
\hline
bc (breast cancer)        & 489             & 105            & 9       & Categorical                      \\
voice     & 2217            & 476            & 20      & Continuous                       \\
iris      & 105             & 23             & 4       & Continuous                       \\
sonar     & 145             & 32             & 60      & Continuous                       \\
soybean   & 478             & 103            & 35      & Categorical, ordinal             \\
vehicle   & 592             & 127            & 18      & Continuous                       \\
vowel     & 693             & 149            & 9       & Continuous                       \\
zoo       & 70              & 16             & 16      & Categorical, continuous          \\
glass     & 304             & 66             & 16      & Continuous                       \\
housevote (house voting) & 304             & 66             & 16      & Categorical                      \\
servo     & 116             & 26             & 4       & Categorical                      \\
ozone     & 256             & 55             & 12      & Ordinal, continuous              \\
boston (Boston housing)   & 354             & 76             & 13      & Categorical, ordinal, continuous	  \\
\bottomrule
\end{tabular}
}
\end{table}

Most of the original datasets have no missing entries. We simulated the missingness by using Algorithm \ref{alg:miss-strategy} to generate two types of missing data: missing not at random with uniform missingness (mnar-uniform) and missing not at random with random missingness (mnar-random). In uniform setting, data records are cropped entirely, i.e. all features are not presenting in missing records. In random setting, data records only miss part of their features. In reality, it won't make much sense to predict a record if there is no available information of it. As a result, most of our work focus on mnar-random case, where we own partial knowledge of the missing records. Fixing $t$ will produce consistent missing rates for different datasets. All the missing entries are filled in by the mean of remained training data on a feature basis, except in last experiment where we investigated the influence of various initialization methods. This means the cropped entries will not be accessed during the whole training and inference process, so the models will not know the ground truth.  The final evaluation metric that we use is the sum of root mean squared error given in following form $\text{RMSE}_{\text{Sum}}(p, t) = \sum_{j=1}^{m} \sqrt[]{\frac{\sum_{i=1}^{n}(p_{ij} - t_{ij})^2 }{n} }$ where suppose p is the imputation result and t is true data matrix, with n sample rows and m feature columns, we compute the root mean squared errors feature-wise and then sum them up.

\begin{algorithm*}[t]
\caption{Algorithm for generating missing entries}
\label{alg:miss-strategy}
\begin{algorithmic} 
\STATE Set target missing percentage $T$ and also state the missing strategy (random or uniform missing). Suppose number of data sample is $N$, randomly generate vector $v$ with size $N$. Randomly choose two attributes $a_1, a_2$ in the dataset
\STATE $t = T$
\WHILE{True}
\IF{uniform} 
    \STATE Set sample attributes to missing where $v < t$ and ($a_1 < \text{mean}(a_1)$ or $a_2 < \text{mean}(a_2)$)
\ELSE
    \STATE Randomly set attributes to missing where $v < t$ and ($a_1 < \text{mean}(a_1)$ or $a_2 < \text{mean}(a_2)$)
\ENDIF
\STATE Calculate the missing percentage $p$ and adjust $t$ if $p \neq T$, otherwise quit the loop
\ENDWHILE
\end{algorithmic}
\end{algorithm*}

\begin{table}[t]
\centering
\caption{Performance on various benchmark datasets with \textbf{50}\% missing rate, generated using the \textbf{mnar-random} strategy.}
\label{tab:rnd-50}
\resizebox{\textwidth}{!}{
\begin{tabular}{l|cccccccccccc}
\toprule
          & Our method         & MICE & Mean & Zero & Medain        & AE   & DAE  & RAE  & MissForest   & KNN  & PCA  & SoftImpute \\ \hline
bc        & \textbf{2.71} & 2.8  & 3.24 & 4.44 & 2.76          & 3.25 & 3.04 & 3.23 & 3.12         & 2.94 & 3.03 & 4.5        \\
voice     & 5.75          & 6    & 6.26 & 15   & 6.48          & 5.81 & 5.8  & 6.24 & \textbf{5.6} & 6.05 & 6.11 & 17.9       \\
iris      & \textbf{0.98} & 1.02 & 1.05 & 3.18 & 1.08          & 1    & 1.02 & 1.04 & 1.07         & 1.1  & 1.05 & 3.02       \\
sonar     & \textbf{16.7} & 22.5 & 16.9 & 34.6 & 16.8          & 17.2 & 17.1 & 17   & 17.1         & 17.5 & 17.1 & 36.9       \\
soybean   & \textbf{7.07} & 9.18 & 7.15 & 25.6 & 7.84          & 7.28 & 7.23 & 7.15 & 7.82         & 8.04 & 7.07 & 23.7       \\
vehicle   & \textbf{2.81} & 4.02 & 2.91 & 13.2 & 2.93          & 2.9  & 2.87 & 2.91 & 2.92         & 3.03 & 2.88 & 12.3       \\
vowel     & \textbf{5.19} & 5.68 & 5.22 & 6.34 & 5.27          & 5.22 & \textbf{5.19} & 5.22 & 5.43         & 5.59 & 5.2  & 8.62       \\
zoo       & \textbf{7.94} & 10.7 & 8.22 & 11   & 9.18          & 8.32 & 8.25 & 8.28 & 8.36         & 8.18 & 8.32 & 15         \\
glass     & \textbf{1.76} & 3.92 & 1.78 & 6    & 1.87          & 1.81 & 1.79 & 1.81 & 1.78         & 1.95 & 1.86 & 7.229      \\
housevote & \textbf{3.1}  & 3.12 & 3.19 & 11.2 & 3.9           & 3.19 & 3.17 & 3.19 & 3.25         & 3.39 & \textbf{3.1}  & 8.8        \\ 
servo    & \textbf{1.15} & 1.32 & 1.22 & 2.89 & 1.23          & 1.26 & 1.2  & 1.2  & 1.32         & 1.28 & 1.32 & 2.67       \\
ozone    & \textbf{3.2}  & 4.53 & 3.27 & 9.14 & 3.31          & 3.28 & 3.25 & 3.22 & 3.48         & 3.47 & 3.36 & 9.94       \\
boston   & 4.65          & 5.03 & 4.34 & 8.39 & \textbf{4.27} & 4.74 & 4.7  & 4.71 & 5.42         & 5.11 & 4.56 & 11.7       \\
\bottomrule
\end{tabular}
}
\end{table}

\begin{table}[t]
\centering
\caption{Imputation error on various benchmark datasets with \textbf{25}\% and \textbf{75}\% missingness, using \textbf{mnar-random} missing generation strategy.}
\label{tab:rnd-vary}
\resizebox{\textwidth}{!}{
\begin{tabular}{l|cccccccccccc}
\toprule
25\% & Our method          & MICE  & Mean         & Zero & Medain       & AE    & DAE           & RAE   & MissForest & KNN   & PCA   & SoftImpute \\
\hline
bc        & \textbf{2.73}  & 3.87  & 2.89         & 3.93 & 2.9          & 3.1   & 3.01          & 2.87  & 3.46       & 3.02  & 2.91  & 4.9        \\
voice     & \textbf{4.35}  & 5.41  & 4.48         & 11   & 4.58         & 4.42  & 4.37          & 4.47  & 4.92       & 4.92  & 4.77  & 14         \\
iris      & \textbf{0.502} & 1.16  & 0.51         & 2.07 & 0.521        & 0.514 & 0.527         & 0.509 & 0.519      & 0.539 & 0.518 & 2.04       \\
sonar     & \textbf{11.4}  & 13.1  & 11.7         & 26   & 12.3         & 11.8  & 11.5          & 11.6  & 11.6       & 12.8  & 11.7  & 35.5       \\
soybean   & 4.3            & 6.91  & \textbf{4.2} & 15.8 & 4.94         & 4.5   & 4.42          & 4.53  & 5.2        & 5.09  & 4.4   & 15.6       \\
vehicle   & \textbf{2.09}  & 4.09  & 2.17         & 10.8 & 2.23         & 2.14  & 2.12          & 2.16  & 2.16       & 2.22  & 2.19  & 10.8       \\
vowel     & 3.43           & 3.72  & 3.48         & 4.33 & 3.54         & 3.48  & \textbf{3.34} & 3.48  & 3.57       & 3.63  & 3.38  & 7.79       \\
zoo       & 2.26           & 4.01  & 2.24         & 2.63 & \textbf{1.6} & 2.29  & 2.14          & 2.24  & 2.65       & 2.06  & 2.38  & 11.6       \\
glass     & 1.76           & 6.63  & 1.74         & 5.29 & 1.88         & 1.74  & \textbf{1.73} & 1.74  & 1.79       & 1.83  & 1.76  & 6.14       \\
housevote & \textbf{1.76}  & 2.1   & 1.78         & 6.94 & 2.46         & 1.79  & 1.83          & 1.78  & 2.02       & 1.95  & 1.78  & 6.39       \\
servo    & \textbf{0.82}  & 0.935 & 0.872        & 1.93 & 0.904        & 0.923 & 0.879         & 0.89  & 0.872      & 0.898 & 0.852 & 2.11       \\
ozone    & \textbf{2.33}  & 3.97  & 2.38         & 6.05 & 2.45         & 2.52  & 2.5           & 2.38  & 2.64       & 2.58  & 2.42  & 7.89       \\
boston   & 1.98           & 3.3   & 2.02         & 4.72 & 2.04         & 1.95  & \textbf{1.93} & 2.01  & 2.36       & 2.37  & 2.04  & 8.27       \\
\bottomrule
\end{tabular}
}
\resizebox{\textwidth}{!}{
\begin{tabular}{l|cccccccccccc}
\toprule
75\%      & Our method         & MICE & Mean          & Zero & Medain & AE   & DAE           & RAE           & MissForest & KNN  & PCA            & SoftImpute \\
\hline
bc        & \textbf{4.15} & 4.94 & 4.7           & 5.95 & 4.2    & 4.77 & 4.73          & 4.71          & 4.97       & 5.13 & 4.87           & 6.33       \\
voice     & 6.91          & 9.57 & 6.97          & 19.3 & 7.06   & 6.88 & \textbf{6.85} & 6.97          & 6.98       & 8    & 6.88           & 20.4       \\
iris      & 1             & 1.63 & 1.05          & 3.88 & 1.06   & 1.05 & 1.03          & 1.05          & 0.97       & 1    & \textbf{0.949} & 3.61       \\
sonar     & 23.7          & 33.9 & \textbf{23.6} & 53.1 & 23.7   & 23.9 & 23.7          & 23.7          & 24.3       & 24.3 & 24.2           & 59.6       \\
soybean   & 9.12          & 12.5 & 9.1           & 30.9 & 9.46   & 9.15 & 9.19          & \textbf{9.09} & 9.68       & 9.73 & 9.15           & 29.5       \\
vehicle   & \textbf{3.5}  & 4.83 & 3.56          & 18.8 & 3.59   & 3.53 & 3.52          & 3.56          & 3.53       & 3.56 & 3.57           & 17.4       \\
vowel     & \textbf{7.16} & 7.99 & 7.42          & 8.92 & 7.58   & 7.43 & 7.34          & 7.33          & 8.07       & 7.8  & 7.34           & 10.1       \\
zoo       & 11.8          & 15.5 & \textbf{11.7} & 14.8 & \textbf{11.7}   & 11.8 & \textbf{11.7}          & 12.2          & 12.1       & 12.1 & 13.2           & 17.1       \\
glass     & \textbf{2.78} & 5.24 & 2.84          & 8.26 & 3.03   & 3.1  & 3.01          & 2.82          & 3.02       & 3    & 2.83           & 9.35       \\
housevote & \textbf{4.24} & 4.38 & 4.34          & 16   & 4.8    & 4.31 & 4.28          & 4.33          & 4.43       & 4.49 & 4.27           & 12.9       \\
servo    & \textbf{1.49} & 1.68 & 1.6           & 3.74 & 1.58   & 1.58 & 1.53          & 1.6           & 1.64       & 1.64 & 1.6            & 3.97       \\
ozone    & \textbf{4.13} & 6.35 & 4.2           & 11.4 & 4.24   & 4.19 & 4.19          & 4.19          & 4.31       & 4.42 & 4.21           & 11.3       \\
boston   & 5.17          & 6.26 & 5.43          & 11.2 & 5.65   & 5.33 & 5.29          & 5.43          & 5.33       & 5.06 & \textbf{4.99}  & 13.9       \\
\bottomrule
\end{tabular}
}
\end{table}

\begin{table}[t]
\centering
\caption{Imputation error on various benchmark datasets with \textbf{50}\% missingness, using \textbf{mnar-uniform} missing generation strategy.}
\label{tab:uniform-50}
\resizebox{\textwidth}{!}{
\begin{tabular}{l|cccccccccccc}
\toprule
50\%      & Our method          & MICE & Mean          & Zero & Medain & AE    & DAE   & RAE   & MissForest    & KNN   & PCA           & SoftImpute \\
\hline
bc        & \textbf{3.85}  & 4.35 & 4.05          & 5.67 & 4      & 4.25  & 4.22  & 4.04  & 4.5           & 4.25  & 4.07          & 6.28       \\
voice     & 5.97           & 6.85 & \textbf{5.96} & 15.1 & 6.06   & 5.99  & 5.97  & 5.97  & 6.09          & 6.48  & 5.97          & 17.5       \\
iris      & \textbf{0.787} & 1.3  & 0.794         & 3    & 0.817  & 0.794 & 0.816 & 0.794 & 0.808         & 0.824 & 0.788         & 3.46       \\
sonar     & \textbf{21}    & 26.2 & 21.2          & 45.3 & 22.3   & 21.6  & 21.2  & 21.1  & 21.9          & 23.1  & 21.3          & 51.2       \\
soybean   & 7.75           & 12.7 & \textbf{7.69} & 26   & 8.59   & 7.81  & 7.8   & 7.69  & 8.77          & 8.55  & 7.78          & 25.8       \\
vehicle   & \textbf{3.07}  & 8.93 & 3.08          & 14.8 & 3.15   & 3.08  & 3.07  & 3.08  & 3.11          & 3.14  & \textbf{3.07} & 16.2       \\
vowel     & 5.39           & 5.95 & 5.43          & 6.95 & 5.53   & 5.45  & 5.41  & 5.43  & \textbf{5.36} & 5.55  & 5.39          & 9.39       \\
zoo       & \textbf{7.09}  & 8.48 & 7.13          & 8.69 & 7.39   & 7.2   & 7.2   & 7.12  & 7.5           & 8.05  & 7.37          & 13.3       \\
glass     & \textbf{2.01}  & 7.12 & 2.03          & 6.4  & 2.18   & 2.05  & 2.04  & 2.04  & 2.15          & 2.12  & 2.07          & 6.69       \\
housevote & 2.72           & 3.74 & \textbf{2.71} & 9.86 & 3.42   & 2.78  & 2.81  & \textbf{2.71}  & 2.87          & 2.92  & 2.75          & 9.05       \\
servo     & \textbf{1.04}  & 1.16 & 1.12          & 2.55 & 1.11   & 1.14  & 1.12  & 1.05  & 1.27          & 1.2   & 1.08          & 2.73       \\
ozone     & \textbf{3.7}   & 7.71 & 3.73          & 9.47 & 3.79   & 3.77  & 3.75  & 3.77  & 4.02          & 3.84  & 3.76          & 9.72       \\
boston    & \textbf{3.66}  & 4.76 & 3.71          & 8.44 & 3.98   & 3.72  & 3.75  & 3.75  & 4.01          & 3.99  & 3.71          & 10.5       \\
\bottomrule
\end{tabular}
}
\end{table}

\begin{table}
\centering
\caption{Classification/ Regression performance on imputed data with \textbf{50}\% missingness, using \textbf{mnar-random} missing generation strategy. The first group datasets is to perform classification task and the accuracy is reported so the higher the value the better the performance, while the second group (name is italic) is for regression task and root mean squared error is reported so the lower the value the better the performance.}
\label{tab:quality-perform}
\resizebox{\textwidth}{!}{
\begin{tabular}{l|cccccccccccc}
\toprule
50\%            & Our method         & MICE          & Mean          & Zero & Medain & AE   & DAE  & RAE           & MissForest & KNN  & PCA  & SoftImpute \\
\hline
bc              & \textbf{93.6} & 83.8          & 88.6          & 87.8 & 87.6   & 92.3 & 92.8 & 93.3          & 87.1       & 87.1 & 87.4 & 87         \\
voice           & \textbf{87.9} & 71.4          & 76.2          & 73.9 & 73.3   & 82.1 & 87.4 & 87.7          & 74.7       & 73.6 & 73.2 & 54.4       \\
iris            & \textbf{61.3} & 59.3          & 51.3          & 47.3 & 47.3   & 54   & 59.3 & 60            & 48.7       & 44.7 & 48.7 & 50         \\
sonar           & 71.1          & \textbf{72.1} & 68.8          & 68.3 & 69.7   & 59.6 & 65.4 & 65.4          & 64.9       & 68.3 & 64.4 & 49.5       \\
soybean         & \textbf{54.9} & 30.7          & 53.9          & 48.2 & 49.3   & 40.1 & 51.4 & 52            & 48.3       & 48.2 & 49.3 & 6.9        \\
vehicle         & 33.8          & 10.1          & 35.2          & 22.3 & 28.6   & 22.5 & 32.2 & \textbf{39.4} & 31.3       & 29.9 & 32   & 5.2        \\
vowel           & 27            & 22.3          & \textbf{30.4} & 26.4 & 25.9   & 4.4  & 21.8 & 22.7          & 29.7       & 27.4 & 26   & 11.2       \\
zoo             & \textbf{61.4} & 41.7          & 50.6          & 41.7 & 41.7   & 56.5 & 57.4 & 58.4          & 52.5       & 50.5 & 50.6 & 37.6       \\
glass           & \textbf{34.1} & 11.2          & 28.6          & 23.4 & 21.9   & 19.1 & 30   & 33.7          & 24.3       & 24.3 & 26.7 & 14         \\
housevote       & 87.4          & 76.3          & 80.7          & 77.5 & 78.9   & 82.8 & 87.6 & \textbf{88.5} & 76.1       & 78.4 & 78.6 & 75.2       \\
\hline
\textit{servo}  & 1.33          & 1.44          & \textbf{1.14} & 1.36 & 1.24   & 1.42 & 1.41 & 1.36          & 1.53       & 1.46 & 1.35 & 1.56       \\
\textit{ozone}  & \textbf{5.59} & 7.9           & 6.13          & 6.52 & 6.28   & 8.59 & 5.74 & 5.69          & 6.24       & 6.36 & 6.29 & 7.31       \\
\textit{boston} & \textbf{6.69} & 8.37          & 7.03          & 7.42 & 7.88   & 7.85 & 7.11 & 6.72          & 7.37       & 7.59 & 7.47 & 8.66       \\
\bottomrule
\end{tabular}
}
\end{table}

\subsection{Experimental Settings}
We used the Adam \cite{DBLP:journals/corr/KingmaB14} optimizer in our experiment. The learning rate was set to 1e-3 and the batch size was set to 50 for all datasets. 
The dimension of the hidden variable $\mb{z}$ was set to 100. We parameterized with feedforward networks having two hidden layers where the number of units is 128 and 64 respectively. For, the networks have two hidden layers as well, with 64 and 128 units respectively. The activation function was set to tanh. In the stochastic gradient variational Bayes algorWe sample from $\mathcal{N}(0, 0.01)$ Gaussian distribution in sample layer. The variance of the Gaussian distribution has an impact on the performance. From the experiment finding it appears that usually the higher the variance the worse performance the model will achieve. We have two extra hyperparameters which are the number of inference operation and number of training step before we perform inference update on training data. They are important for the model performance. Too much inference in early training stage will not help the imputation as the model parameters are still not well updated. In most of our experiments the inference time is set to 2 and the training interval is set to 5. And we run 100 epoch to convergence.

We compared with the following 11 baselines: Mean/Median imputation, Zero imputation, SoftImpute~\cite{mazumder2010spectral}, MICE~\cite{buuren2010mice}, MissForest~\cite{stekhoven2011missforest}, KNN~\cite{altman1992introduction}, PCA, AE~\cite{rumelhart1985learning}, DAE~\cite{DBLP:journals/corr/GondaraW17}, and RAE~\cite{Tran2017MissingMI}. 
For Mean/Median imputation, the missing entries are filled by the feature-wise mean/median value computed from observed data, and Zero imputation fills missing entries by zero.
SoftImpute uses iterative soft-thresholded singular value decomposition to impute the missing value.
MICE iteratively imputes each feature by regression on the rest features.
MissForest is a non-parametric approach that directly predicts missing values by training a random forest model on observed data.
KNN approximates the missing value in each data sample by the corresponding value of the closest data example.
PCA fits on the incomplete data with some initialization on the missing entries and estimates them with calculated principal components.
AE, DAE, and RAE work mostly in the same manner. They all have an encoder to map the input to some encoded feature and use a decoder to map the feature to some output. For AE the input and output are the same. For DAE the input is the incomplete data with added noise on observed entries and the output is the original incomplete data. For RAE the input is the same as DAE one but the output is the residual between the corrupted and original incomplete data.
All three models share the same model structure as our method uses. Encoded dimension can be tuned to achieve better solution for each model. On each dataset, each model is run multiple times to get multiple imputation result for analysis so we take advantage of Multiple Imputation (MI) \cite{schafer1999multiple}.

Besides these deep models, we also compare our method with MICE, MissForest, KNN, PCA and SoftImpute. For these methods we use existing packages to record the final performance. The error analysis uses $ \text{RMSE}_{\text{Sum}} $ as evaluation metric performed on scaled data in order to remove influence from diverse ranges of features.

\subsection{Experiment Results}
Table \ref{tab:rnd-50} shows the imputation performance on datasets with a 50\% missing rate. Our method outperforms all the baselines on 11 out of 13 datasets by reducing error up to \textasciitilde 7\%. This shows the generative model with EM algorithm can lead the generated missing value closer to the ground truth, i.e. $\hat{p}(\mc{Y} \mid \mc{X})$ closer to true $p(\mc{Y} \mid \mc{X})$, in various situations. The Mean/Median/Zero imputation only use fixed numbers to replace the missing values which cannot fully represent the intrinsic correlation among features and only serve as a general and blur estimation, although mean and median remain strong competitors in our compared baselines, which is due to the reason that when the data is not skewed, i.e. there doesn't exist much outliers in the dataset, the mean/median can represent the most common information of the data. MICE, KNN and SoftImpute are inherently linear and cannot capture nonlinear feature patterns so they perform poorer than our method. Among MICE, MissForest, KNN, PCA and SoftImpute, MissForest takes longer time to train and estimate the missing value since it requires quite amount of trees to make decision on complicated dataset. MissForest and KNN can easily overfit on training data. And since they are based on heuristic rules, they don't capture inter-correlation among features and perform poorer than our method. Such similar performance is also shown in our qualitative analysis, which will be discussed later.

Table \ref{tab:rnd-vary} shows the imputation performance on datasets with various missing rates, 25\% and 75\% in our setting to represent low and high missing situation. When the rate is low, our method outperforms baselines on 8/13 datasets while most of the methods tend to achieve comparable results. It is intuitive as when the missing rate is small, the observed data can provide enough information to infer the missing entries using different models. When the rate is super high, it is hard for models to learn inter-correlation among features. In such situation the imputation raises a higher demand on model to capture those relations. The performance is more diverged compared to previous results while our method still outperforms other baseline methods on over a half cases. In reality, when the dataset is largely missing, it is not wise to directly impute the missing data. It is better to use techniques such as heuristics or correlation analysis to remove unnecessary features in order to reduce the missing rate and then perform the missing imputation task. And although not a main point of study, we still report the performance on a setting of 50\% uniformly missing data, as show in table \ref{tab:uniform-50}, in which our method still achieves better results on most datasets. All above results demonstrate the power of our model to learn data pattern and generate more real data for missing values.

The main goal of imputing missing value is to generate real-like data to help with analysis of the dataset, for example, train a machine learning model to classify an unseen data point. Apparently in general, with more information in data, the classification will be more accurate. So besides the analysis of how close the imputed data is to the true data, we also care about how meaningful the imputed data is to the analytics, i.e. how much intrinsic correlation among features we catch in the imputed data. We fit multi-layer perceptron (MLP) model for all datasets and change the loss between cross entropy loss and squared error loss based on the task. For the classification task we use cross entropy loss and report test accuracy and for the regression task we use squared error loss and report final root mean squared error (RMSE) as evaluation metric.

\begin{wraptable}{r}{0.4\linewidth}
\caption{Imputation error on various benchmark datasets with different initialization methods, under \textbf{50}\% missingness, using \textbf{mnar-random} missing generation strategy.}
\label{tab:init}
\resizebox{\linewidth}{!}{
\begin{tabular}{l|cccc}
\toprule
          & Mean          & Zero          & Median       & Random \\
\hline
bc        & 2.71          & 2.73          & \textbf{2.6} & 3.33   \\
voice     & \textbf{5.75} & 6.17          & 5.94         & 10.6   \\
iris      & \textbf{0.98} & 2.04          & 1.03         & 2.49   \\
sonar     & \textbf{16.7} & 25.2          & 26.6         & 65.1   \\
soybean   & \textbf{7.07} & 11.6          & 7.86         & 14     \\
vehicle   & \textbf{2.81} & 5.39          & 3.08         & 6.2    \\
vowel     & 5.19          & \textbf{5.17} & 5.2          & 5.73   \\
zoo       & \textbf{7.94} & 10.1          & 9.33         & 11.4   \\
glass     & \textbf{1.76} & 4.68          & 3.42         & 8.37   \\
housevote & \textbf{3.1}  & 6.8           & 3.12         & 7.26   \\
servo     & \textbf{1.15} & 2.03          & 1.2          & 2.06   \\
ozone     & \textbf{3.2}  & 5.37          & 3.94         & 4.82   \\
boston    & \textbf{4.65} & 7.23          & 6.22         & 9.15   \\
\bottomrule
\end{tabular}
}
\end{wraptable}

Table \ref{tab:quality-perform} shows the results performance of classification/regression tasks on datasets with 50\% missing rate using mnar-random missing generation strategy. The Mean remains a strong competitor besides RAE model. The result, in a qualitative way, demonstrates that our method better captures the intrinsic data pattern on a wide variation of datasets. And the imputed data possess higher quality in different data analysis tasks.

Lastly we investigate the model performance given different initialization. We compare using mean/zero/median and random value to initialize the dataset. The random value is generated with a $\mathcal{N}(0, 1)$ Gaussian distribution. The result is shown in table \ref{tab:init}. We see initializing with mean value yields the best performance on most datasets. This makes sense as we iteratively use imputed data to train the model and update the missing entries with trained weights, it is better to start from a good indicator. A deviated starting point will make the initial learning of our method hard to capture real correlation among features and thus make the entire process not able to achieve best imputation result.

\section{Conclusions and Future Works}
In this paper, we study the imputation of missing values. We formulate this problem as a generative process: give the observed values $\mc{X}$, generate the missing ones $\mc{Y}$, by estimating a conditional distribution $p(\mc{Y}|\mc{X})$. We show that this distribution can be estimated by solving a fixed-point problem defined on two conditional distributions that are parameterized by deep neural networks. In experiments on 10 datasets, our method outperforms 11 baseline methods.

For future works, we will extend our method to missing value imputation in time-series data. The general probabilistic framework developed in Section~\ref{sec:method} can be largely re-used. The two conditional distributions need to be redesigned to accommodate the time-series structure. Possible candidates are the variational recurrent network~\cite{chung2015recurrent}.

Another direction to explore is task-aware missing value imputation. Instead of performing the MVI and the target task sequentially: first filling the missing data, then feed the imputed data to the downstream task, we can perform them jointly to better explore the synergistic relationship between them. On one hand, more accurate imputation of missing values prepares better inputs for the target task, which boosts
the task' performance. On the other hand, during training, the performance of the target task provides guidance to MVI such that the imputed values are more suitable for this task. 

\bibliographystyle{plain}
\bibliography{ref}

\end{document}